\documentclass[a4paper,twoside]{article}

\usepackage{epsfig}
\usepackage{subcaption}
\usepackage{calc}
\usepackage{amssymb}
\usepackage{amstext}
\usepackage{amsmath}
\usepackage{amsthm}
\usepackage{multicol}
\usepackage{pslatex}
\usepackage{apalike}
\usepackage[bottom]{footmisc}

\usepackage{booktabs}
\newcommand{\normal}{\textit{normal}}
\newcommand{\sensor}{\textit{sensor}}
\newcommand{\clean}{\textit{clean}}

\usepackage{array}
\usepackage{tabularx}
\usepackage{multirow}
\newcolumntype{Y}{>{\centering\arraybackslash}X}
\newcolumntype{Z}{>{\centering\let\newline\\\arraybackslash\hspace{0pt}}X}
\newcolumntype{R}{>{\raggedleft\arraybackslash}X}

\usepackage{SCITEPRESS} 

\begin{document}
\pagenumbering{gobble}
\title{Let's Get the FACS Straight - Reconstructing Obstructed Facial Features}
\author{
  \authorname{Redacted for Peer-Review}
}
\author{
  \authorname{
    Tim B\"uchner\sup{1}\orcidAuthor{0000-0002-6879-552X}, 
    Sven Sickert\sup{1}\orcidAuthor{0000-0002-7795-3905}, 
    Gerd Fabian Volk\sup{3}\orcidAuthor{0000-0003-1245-6331}, 
    Christoph Anders\sup{2}\orcidAuthor{0000-0002-5580-5338}, 
    Orlando Guntinas-Lichius\sup{3}\orcidAuthor{0000-0001-9671-0784} and 
    Joachim Denzler\sup{1}\orcidAuthor{0000-0002-3193-3300},
  }
  \affiliation{\sup{1} Computer Vision Group, Friedrich Schiller University Jena, Jena, Germany}
  \affiliation{\sup{2} Division of Motor Research, Pathophysiology and Biomechanics, Clinic for Trauma, Hand and Reconstructive Surgery, University Hospital Jena, Jena, Germany}
  \affiliation{\sup{3} Department of Otolaryngology, University Hospital Jena, Jena, Germany}
  \email{\{tim.buechner, sven.sickert, joachim.denzler\}@uni-jena.de, \{fabian.volk, christoph.anders, orlando.guntinas\}@med.uni-jena.de}
}
\abstract{
  The human face is one of the most crucial parts in interhuman communication.
  Even when parts of the face are hidden or obstructed the underlying facial movements can be understood.
  Machine learning approaches often fail in that regard due to the complexity of the facial structures.
  To alleviate this problem a common approach is to fine-tune a model for such a specific application.
  However, this is computational intensive and might have to be repeated for each desired analysis task.
  In this paper, we propose to reconstruct obstructed facial parts to avoid the task of repeated fine-tuning.
  As a result, existing facial analysis methods can be used without further changes with respect to the data.
  In our approach, the restoration of facial features is interpreted as a style transfer task between different recording setups.
  By using the CycleGAN architecture the requirement of matched pairs, which is often hard to fullfill, can be eliminated.
  To proof the viability of our approach, we compare our reconstructions with real unobstructed recordings.
  We created a novel data set in which 36 test subjects were recorded both with and without 62 surface electromyography sensors attached to their faces.
  In our evaluation, we feature typical facial analysis tasks, like the computation of Facial Action Units and the detection of emotions.
  To further assess the quality of the restoration, we also compare perceptional distances.
  We can show, that scores similar to the videos without obstructing sensors can be achieved.
}
\onecolumn \maketitle \normalsize \setcounter{footnote}{0} \vfill
\section{\uppercase{Introduction}}
\label{sec:introduction}
Assessing a human's emotional state by means of facial expressions is an ability which required humanity over millions of years to learn.
Researchers are interested in the automatic classification of these expressions based on input signals (images, videos, locally attached sensors, etc.) to infer the connected underlying emotional states.
The continuos progress in machine learning, especially with respect to computer vision tasks, significantly improved the classification accuracy~\cite{luanresmaskingnet2020}.
Often such models are used in medical and psychological studies, or in-the-wild applications whereas their emotional assessment is debateable~\cite{barrettWasDarwinWrong2011,heavenWhyFacesDon2020}.
Furthermore, the connection between facial expressions and the actual underlying mimetic muscles is still an open research question.
The Facial Action Coding System (FACS)~\cite{hjortsjoManFaceMimic1969,ekmanFacialActionCoding1978} was a first approach to build a connection between those two and is still used, today.
However, as FACS is based on the facial landmarks the connections to the muscles are abstracted via proxies.

A dominant problem for machine learning based methods applied in these scenarios is the acquisition and content of the training data set.
The intended and actual usage of these might differ significantly and thus could lead to unreliable or even undesirable results.
For instance, in regards to the classification of facial expressions in FER2013~\cite{goodfellowChallengesRepresentationLearning2013}, obstructions in the face have not been considered.
We show in our work that, instead of fine-tuning models to a custom data set the obstructed features can also be correctly reconstructed.
To demonstrate this, we recorded a custom data set measuring the face and the muscle activity simultaneously.
A single recording contains the following tasks: eleven facial movements~\cite{schaedeVideoInstructionSynchronous2017}, five spoken sentences, and ten mimics of emotions.
To restore the facial features, we created a recording setup in which each test subject was recorded with and without sEMG sensors attached to their face.
\begin{figure*}
  \centering
  \includegraphics[width=\textwidth]{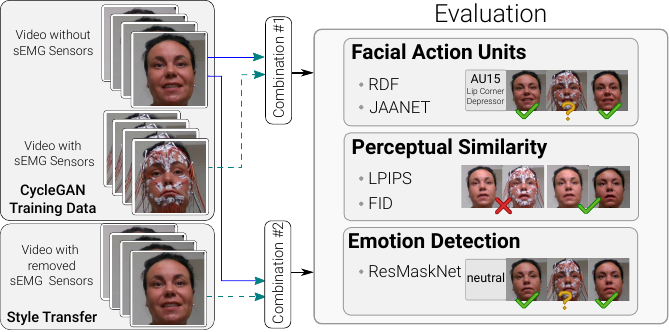}
  \caption{
    Our experimental setup to evaluate the correct restoration of facial features:
    A video without sEMG sensors represents our baseline (\normal).
    For comparison we have videos with those sensors visible (\sensor) and videos where they have been removed by our proposed approach (\clean).
    Our evaluation includes the tasks of extracting Facial Actions Units and emotion detection.
    Furthermore, we analyze their perceptual similarity in comparison to the baseline.
    Green check marks, red crosses and yellow question marks indicate similarity or the possibility to solve a task given the underlying data.
  }
  \label{fig:experimental_setup}
\end{figure*}
In our experimental setup, we show that state-of-the-art algorithms fail to correctly solve their intended task.
A brief overview is depicted in Fig.~\ref{fig:experimental_setup}.
We evaluate the extraction of the Facial Actions Units (AUs) and emotion detection.
For each analysis the video without sEMG sensors represents the baseline, which we aim to achieve in our restoration approach.
Additionally, we assess the visual quality with two perceptual scores.

In order to recover the obstructed facial features, several complex tasks need to be solved simultaneously.
There are 36 individual test subjects in the data set with a large visual variance.
Although an instruction video is given, the timing and intensity of carrying out a task varies a lot, even within the recordings of the same test subject.
Hence, a pair-wise matching between corresponding frames of the videos would be extremely difficult.
Furthermore, the correct facial expression has to be recreated in their correct intensity.
Otherwise the estimation of the AUs is likely to fail.

Due to the complexity of this task, traditional methods like segmentation and image inpainting are not an option.
However, in our setup we are using sEMG sensors to measure muscle activity.
To ensure correct measurement those sensors have to placed at the same anatomical locations each time.
Thus, we propose to represent this strict placement of the sensors as a consistent style change between two images of the same person independent from the facial expression.
In combination with the unmatched-pair setting, we can deploy the CycleGAN architecture by Zhu et al.~~\cite{zhuUnpairedImagetoImageTranslation2017} to learn this style transfer.

This proposed approach retains the visual appearances of the test subjects.
In fact, we show that completely covered facial features can be restored correctly.
With respect to quality, our~\clean~videos resemble the~\normal~videos more than the~\sensor~videos.
More importantly, downstream facial analysis algorithms can be applied directly without the need of fine-tuning them first for images with sEMG sensors.
We eliminate the problem of obstructed facial features that otherwise would render an in-depth analysis of expressions and muscle activity impossible.

\section{\uppercase{Related Work}}
To restore obstructed facial features, we mention in the following related approaches in the area of generative models.
There are approaches that similarly aim to either transfer styles or restore missing features.
However, non of them focus on correctly restoring facial features for further down-stream applications.

The first method for unmatched-pair style transfer was established by Zhu et al.~\cite{zhuUnpairedImagetoImageTranslation2017} with the introduction of CycleGAN.
Their work mostly focuses on the visual stability and quality during the translation task.
The introduced \textit{consistent cycle loss} for reducing the underlying mapping distributions helped with stabilizing training.
In general, CycleGAN can be deployed for different style transfer tasks and is applicable to in-the-wild images.
We propose to use this method to impaint structured obstructions in images including frontal face recordings to restore underlying properties of hidden facial features.
The model must learn an internal representation of the facial movements in order to restore them.

Another specific task with removing and adding facial obstructions can be seen in the transfer of makeup styles between people.
Nguyen et al.~\cite{nguyenLipstickAinEnough2021} proposed a holistic makeup transfer framework.
In their work, they were able to retain facial features, but artificially add light and extreme makeup styles on in-the-wild images.
It shows that even complicated obstruction patterns can be learned by neural networks.
For our data set the problem definition is easier as the sEMG sensors are always at the same location.
However, they cover around $50\%$ of the crucial facial areas and the color of the connected cables varies.

Li et al.~\cite{liGenerativeFaceCompletion2017} use generative adversarial networks to restore randomly altered human faces.
They artificially crop random areas inside the facial bounding box and replace them with noise values.
Their model creates high qualitative visual results.
At the same time, it might create different visual expressions depending on the noise patch.
We have to retain the correct underlying facial structure and expression.
Thus, although not fully suited in our scenario, their approach of using GANs for inpainting missing information serves as a good starting point.

To restore facial features, Mathai et al.~\cite{mathaiDoesGenerativeFace2019} proposed an encoder-decoder architecture with matched pairs.
In their work, they artificially place obstructions like sunglasses, hats, hands, and microphones onto faces to improve the robustness of facial recognition software.
The model learns to replace the underlying facial features using the learned auto-generative capabilities.
However, the work does not account for similarity of facial expressions and movements.
Only the facial similarity for biometric purposes was relevant.

In summary, we can assess that generative models can retain important facial features while enabling correct person recognition.
In our research, we make use of the unmatched-pair property of CycleGANs~\cite{zhuUnpairedImagetoImageTranslation2017} to both attach and detach sEMG sensors.
It is worth noting that we are not only interested in the visual quality of the generated videos.
We also want to ensure that existing state-of-the-art facial analysis methods for down-stream tasks produce correct results.
In our scenario, it would allow us to make use of the simultaneous recordings of sEMG and video signals.

\section{\uppercase{Data Set}}
\label{sec:data}
\begin{figure}[t]
  \centering
  \includegraphics[width=\linewidth]{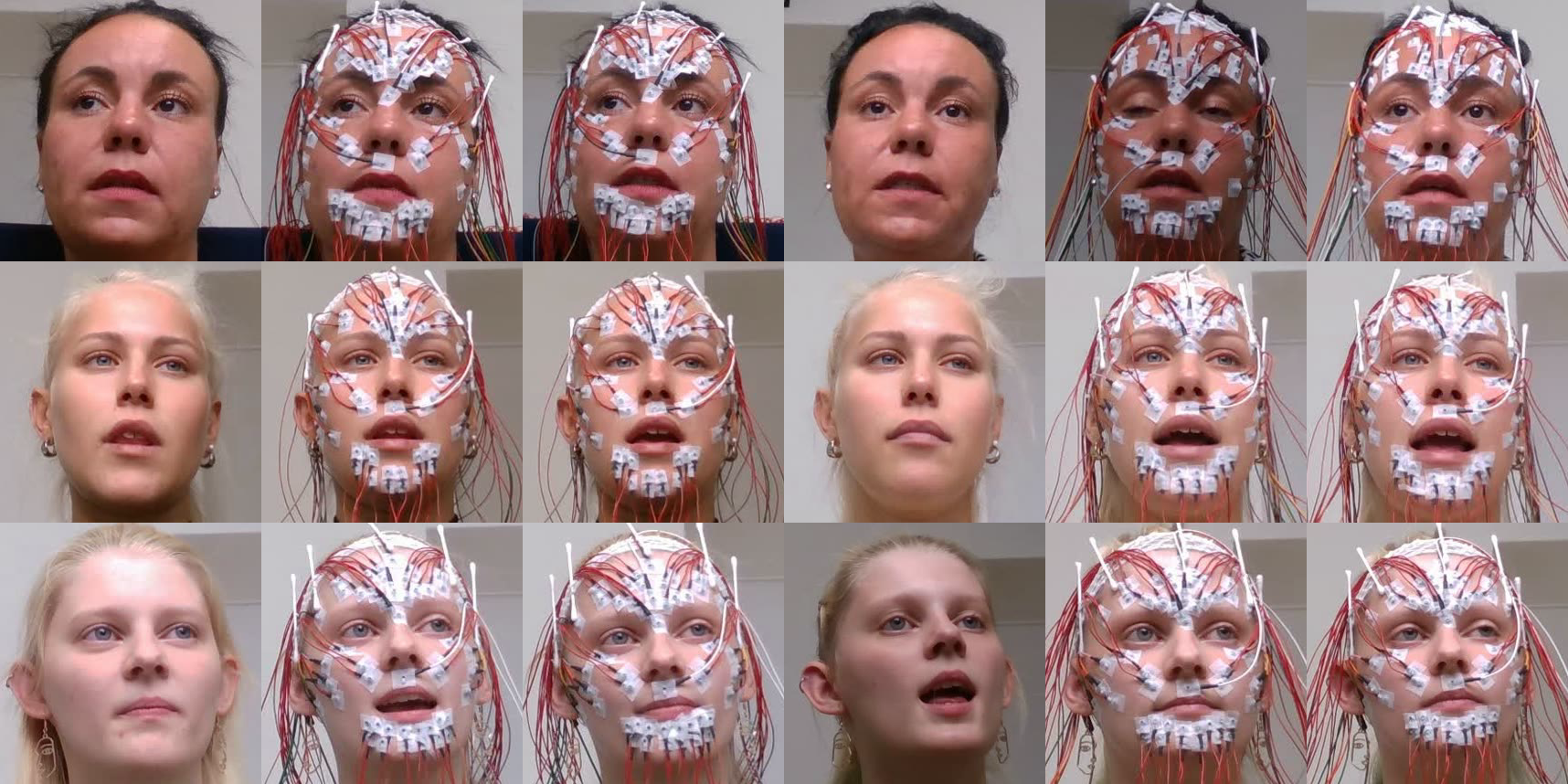}
  \caption{
    Overview of three selected test subjects with their three measurements on each of the two recording dates.
    For each subject one recording without attached sensors and two with attached sensors is displayed.
    The 62 sEMG sensors are attached to the same anatomical locations for all test subjects.
    The sensors block relevant facial areas, such as the forehead, completely.
  }
  \label{fig:dataset:overview}
\end{figure}

In our work we are interested in learning about the connection between mimics and muscles.
Thus, we created a new data set measuring both domains simultaneously.
We recorded the facial movement and muscle activity of 36 test subjects\footnote{All shown individuals agreed to have their images published in terms with the GDPR.}, with 19 identifying as female and 17 identifying as male.
Facial movements were captured using a frontal facing camera with a resolution of $1280 \times 720$ and $30$ frames per second.
Muscle activity was recorded using surface electromyography (sEMG).
For a full measurement we attached 62 sEMG sensors to the face, including connector cables.
Fig.~\ref{fig:dataset:overview} shows the sensor placement of three selected test subjects.
A single sensor consists of a white connection patch on the skin and either a red or white cable.
Furthermore, white cotton swabs were used to fix the cables at their position.
During the attachment of the sensors it is possible that the color order of the cables change among different test subjects.

To be able to learn the restoration of facial features, we recorded subjects once without attached sEMG sensors and twice with attached sEMG sensors.
These recording sessions were repeated again after two weeks under the same conditions.
There was no order, whether subjects were recorded first with attached sensors or without them.
It is also not relevant for the correction of the facial features.
However, for each measurement an instruction video was shown to ensure the same order of tasks to enable a later comparable analysis.

The given instruction tasks can be divided into three subgroups.
In the first task subjects need to mimic eleven distinct facial expressions three times.
We follow the protocol of Schaede et al.~\cite{schaedeVideoInstructionSynchronous2017} and refer to this as the \textit{Schaede} task in the remainder of this paper.
In the second tasks, subjects need to repeat five spoken sentences, which we will refer to as \textit{Sentence}.
The last task is the imitation of 24 shown basic emotional expressions and will be called \textit{Emotion} task.
This split of tasks is intended to cover different areas of activity for the facial muscles.
In total 174 videos were recorded with a ratio of 1:2 \textit{normal} and \textit{sensor} videos, respectively.
To reduce the influence of background noise in the videos, we run our experiments only on the facial areas of the test subjects.
Details about the face extraction can be found in the next section.

Medical experts observed the experiments and manually started and stopped the capturing devices for the video and sEMG recordings.
Hence, a one-to-one frame-wise matching between the \textit{normal} and the \textit{sensor} videos is not possible.
Time delays among the recordings of an individual test subject cannot be estimated.
Even though the test subjects follow a given instruction video, the intensity of the facial expressions can vary significantly.
Further, subjects do not always start the task at the same time.
This might be due to fatigue and the repeated nature of these tasks as a recording session takes around 1.5 hours.

Additionally, the test subjects change their head posture, gazing angle, and distance to the camera throughout all recordings.
Among the different recording sessions the illumination inside the room changes, which in turn results in different appearances of the cables.
Faces might also be obstructed by hands as sometimes detached sEMG sensors had to be reattached during the recording.
These constraints require a method which can work without matched pairs.
Additionally, the method should be adaptive to avoid overfitting to illumination settings and cable colors in the training data.

\section{\uppercase{Methods}}
A lot of machine learning methods learn facial features to solve their respective tasks, like emotion detection, person identification or visual diagnostics.
However, many of these only work on non-obstructed faces and would require fine-tuning.
We aim to restore their capabilities without the need of adapting the models.
For our given problem, we can define the facial obstructions in a structured manner.
The anatomical correction sensor placement ensures that it can be described in such a way, as shown in Fig.~\ref{fig:dataset:overview}.
Thus, we can reinterpret the changes between a~\normal~and~\sensor~video frame as a style transfer for which we have to learn a translation.
In the following, we will refer to the frames and videos in which the sEMG have been removed as~\clean.
We first define the CycleGAN~\cite{zhuUnpairedImagetoImageTranslation2017} architecture with the adaption to our task.
As we are not only interested in the visual appearance of the generated images but also in the restored facial features, we describe methods for perceptual metrics and facial feature comparison.

\subsection{Removal of sEMG Sensor using Unpaired Style Transfer}
As described in Sec.~\ref{sec:data} several challenges arose during the video acquisition.
Among these challenges are the unmatched-pairs of video frames, the subtle changes in the recording environment, and the test subjects head posture, angle, gaze and location changes.
However, the obstruction of the facial features by sEMG sensors occurs in a structured manner.
Thus, a translation model between the~\normal~and~\sensor~video frames could be learned to correctly restore the facial features independent from the underlying facial expression.
\begin{figure}[t]
  \centering
  \includegraphics[width=\linewidth]{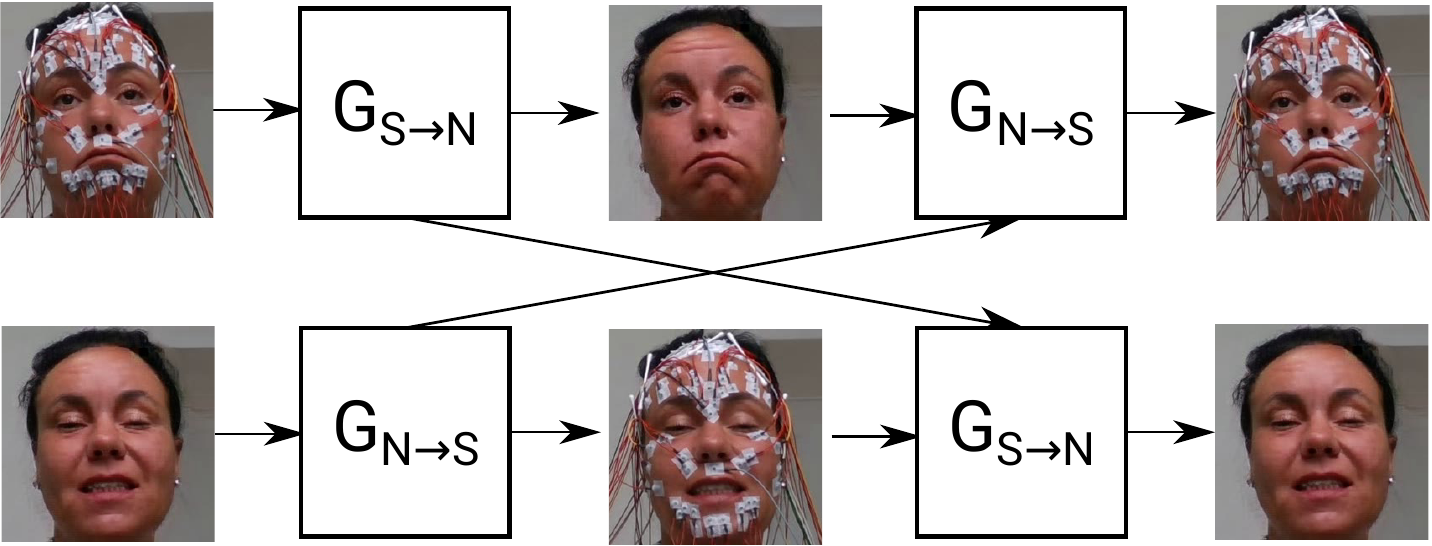}
  \caption{
    Double generative structure of the CycleGAN for the proposed sEMG sensor removal.
    Generator $G_{N \mapsto S}$ learns the attaching of the sensors.
    Generator $G_{S \mapsto N}$ learns the detaching of the sensors.
    Different facial expressions can be combined without being changed during the translation process.
  }
  \label{fig:methods:ganstructure}
\end{figure}

The issue of unmatched-pairs style translation was solved by Zhu et al.~\cite{zhuUnpairedImagetoImageTranslation2017} with the introduction of CycleGANs.
Instead of relying on a single generator-discriminator-architecture~\cite{goodfellowGenerativeAdversarialNetworks2014} they jointly trained two opposing translation generation tasks.
The double generator structure is shown in Fig.~\ref{fig:methods:ganstructure} and we tuned it towards the removal of the sEMG sensors.

The generator $G_{N \mapsto S}$ learns the mapping from the~\normal~domain to the~\sensor~domain, whereas generator $G_{S \mapsto N}$ learns the inverse direction.
For a given unmatched training input pair $(N_{in}, S_{in})$ this architecture computes the corresponding output pair $(S_{out}, N_{out})$.
To ensure stable training Zhu et al. introduced the \textit{consistent cycle loss}~\cite{zhuUnpairedImagetoImageTranslation2017}, the \textit{adversarial loss}~\cite{goodfellowGenerativeAdversarialNetworks2014}, and the \textit{identity loss}~\cite{taigmanUnsupervisedCrossDomainImage2016}.
The discriminator $D_{S}$ uses the~\sensor~image to estimate the generated image by $G_{N \mapsto S}$ to check whether they come from the same source distribution.
The discriminator $D_{N}$ handles the inverse direction similarly.
To ensure the correct restoration of the obstructed facial features in our analysis, we only use the generator $G_{S \mapsto N}$ by translating the~\sensor~video to the~\clean~video version.

\subsection{Feature Restoration Evaluation of Cleaned Videos}
We evaluate the success of the restoration of the obstructed facial features by two means.
In the first evaluation, we assess the visual quality of the generated images by the generator $G_{S \mapsto N}$ by comparing their perceptual similarity to frames of the~\normal~video.
Hereby, we compute two perceptual similarity scores.
We calculate the image-to-image similarity through all frame pairs between the~\normal~and the~\clean~video.
Furthermore, we estimate the underlying image distributions between these videos and compute their similarity.

In the second evaluation, we check the correct restoration of the facial features by applying well-known machine learning algorithms for facial analysis.
Specifically, we evaluate the fitting of Facial Action Units (AUs) and emotion detection restoration.
For AUs, we use random decision forest~\cite{breimanRandomForests2001} and the attention-based JAA-Net model by Shao et al.~\cite{shaoJAANetJointFacial2021}.
Both models have already been implemented in the library PyFeat~\cite{cheongCosanlabPyfeat2022}, which we use for our evaluation.
For emotion detection we use ResMaskNet by Luan et al.~\cite{luanresmaskingnet2020}, which still yields the current state-of-the-art performance.
To further show that our restoration approach produces convincing results, we run the same evaluations also as comparison between the~\normal~and~\sensor~videos.
The whole experimental setup is depicted in Fig.~\ref{fig:experimental_setup} summarizing all comparisons.
We compare the resulting time series with each others using dynamic time warping (DTW)~\cite{lhermitteComparisonTimeSeries2011} and mean absolute percentage error (MAPE).
With DTW we can avoid the unknown delay between each recording, and for MAPE we compute all possible shifts in a time frame of $\pm$ 20 seconds.

\subsubsection*{Learned Perceptual Image Patch Similarity}
Zhang et al.~\cite{zhangUnreasonableEffectivenessDeep2018} introduced the LPIPS score for image-to-image similarity measurements.
They compute the $L_2$ distance between the feature vectors of the last convolutional layer of classification models, either AlexNet~\cite{krizhevskyImageNetClassificationDeep2012} or VGG~\cite{simonyanVeryDeepConvolutional2015}., to estimate perceptual similarity.
Under the premise of similar looking images, which would lead to same classification results, the features vectors should be similar, as well.
The metric ranges from $0$ for image identity, to $1$, indicating no perceptual similarity.

As the sEMG sensors cover a substantial area of the test subjects' faces, we assume that there is a high distance between the frames of the~\normal~and~\sensor~videos.
Furthermore, we assume that the feature vectors of the deep learning models stay similar for a test subject regardless of their facial expression, head movement, or glance.
The pre-training on ImageNet~\cite{dengImageNetLargescaleHierarchical2009} does not include these fine-grained classification tasks and thus should yield the same feature vectors.
If the~\clean~video generated by $G_{S \mapsto N}$ produces a low LPIPS score, the restoration of the facial features might have been already successful.

\subsubsection*{Fr\'echet Inception Distance}
The image-to-image comparison alone might be insufficient, since correct matching between the frames is not possible.
However, comparing the unknown underlying generative distribution of the video frames could lead to a more reliable understanding.
We use the Fr\'echet Inception Distance (FID) introduced by Heusel et al.~\cite{heuselGANsTrainedTwo2017} to compare these generative distributions.
The FID assumes that both source and target are multivariate normal distributions.
Thus, the parameters have to be estimated.
We use Inception v3~\cite{szegedyRethinkingInceptionArchitecture2015} feature vectors to estimate the mean vector and the covariance matrix.
For the actual implementation, we use the FastFID approximation by Mathiasen and Hvilsh{\o}j~\cite{mathiasenBackpropagatingFrechetInception2021}, which has a deviation of $0.1$ to the original distance but is significantly faster.
However, a drawback of the FID is the dependence on the batch size.
To ensure comparable results all evaluations are run with the same batch size of $N=128$.
The problems raised by Liu et al.~\cite{liuImprovedEvaluationFramework2018} do not affect our evaluation as our task is class independent.

\subsection{Implementation Details}
To learn the translation from faces with and without attached sEMG sensors, we use the CycleGAN~\cite{zhuUnpairedImagetoImageTranslation2017} architecture which fits our problem task the best.
In a first ablation study we discovered that a test subject unique model creates better results. 
This way we can mitigate the impact of other possible latent translation tasks the model could pick up like gender or age.
For each subject we acquired six videos, four with and two without sEMG sensors.
To create the training data, we randomly choose frames from the videos.
Further, we limit number of training data to $2\%$ of the available frames to learn the sEMG style-translation model.

We recording setup was defined in such a manner that the test subjects do not move their heads.
Thus, a fixed bounding box location for the videos was defined to ensure same head sizes and margins to the background.
We rescale all extracted faces to a size of $286 \times 286$ to match the backbone generator network.
Then we split the extracted frames into $90\%$ training and $10\%$ validation data.
During the training the images are augmented using horizontal flipping, random cropping, and normalization into the range $[-1, 1]$.
The model evaluation is then done on the remaining full test subject videos.

\begin{figure}[t]
  \centering
  \includegraphics[width=\linewidth]{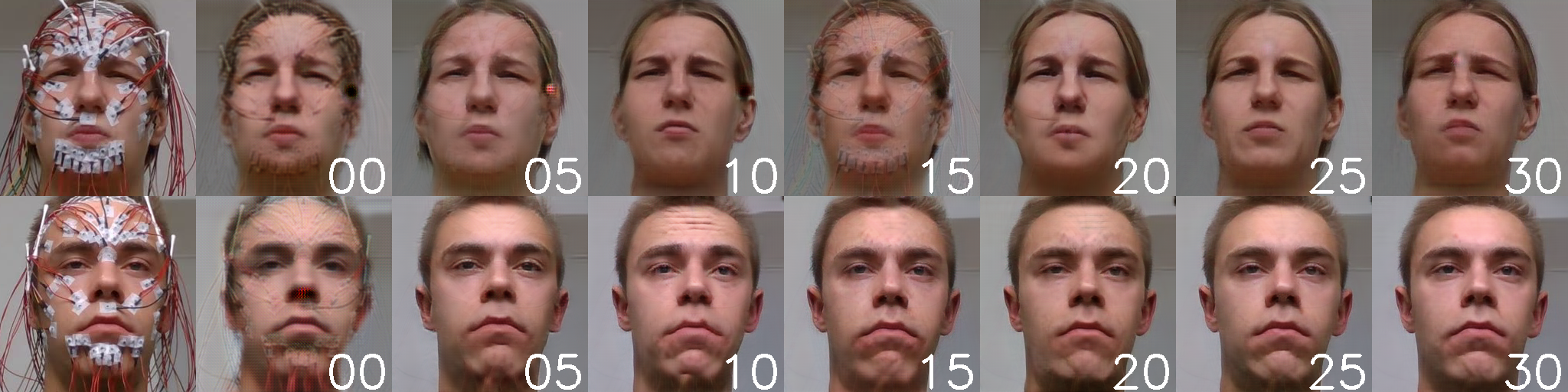}
  \caption{
    We display the trainings progress of the sEMG sensor removal.
    During the first $5$ epochs the model focuses on the general removal of the sensors.
    After that, the more fine-grained details in the faces are restored.
  }
  \label{fig:training_progress}
\end{figure}

We train a ResNet~\cite{heDeepResidualLearning2016} model with nine blocks from scratch as the generator networks.
Before the ResNet blocks two additional down-sampling blocks are added, which are then reversed at the end.
The PatchDiscriminator by Isola et al.~\cite{isolaImagetoImageTranslationConditional2017} builds the foundation for both discriminators.
All models are trained for $30$ epochs with an initial learning rate of $3e^{-4}$ and a continuos linear learning rate decay update after $15$ epochs.
In Fig.~\ref{fig:training_progress} we show the training progress of the $G_{S \mapsto N}$ generator.
It can be seen that the model learns the general removal of the sEMG sensors immediately and then focuses on restoring fine-grained facial details.

\section{\uppercase{Results}}
To validate the correct restoration of the facial features we use the described experimental setup in Fig.~\ref{fig:experimental_setup}.
For each test subject we translate the four~\sensor~videos with the specialized $G_{S \mapsto N}$ generator.
Inside the video $98\%$ of the frames have not been seen by the generator and thus serve as test set.
The perceptual qualitative comparison was done using the LPIPS and FID scores.
Then we also extract the AUs and eight basic emotions.
However, due to the high amount of possible video testing combinations we chose a suitable subset of all these combinations.
We compare only videos from the same recording session with each other.
This reduces possible interference due to background changes or clothing changes of the test subjects.
For all results, we investigate the three given tasks: Schaede, Sentence, and Emotion.
As a baseline for comparison we compute all evaluations between the two~\normal~videos ($N_{1}$, $N_{2}$) recorded at each of the two sessions.
In all tables and figures the sessions will be indicated by a subscript.
We assume that the similarities between these session are limited to ensure comparability.

\begin{figure}[t]
  \centering
  \includegraphics[width=\linewidth]{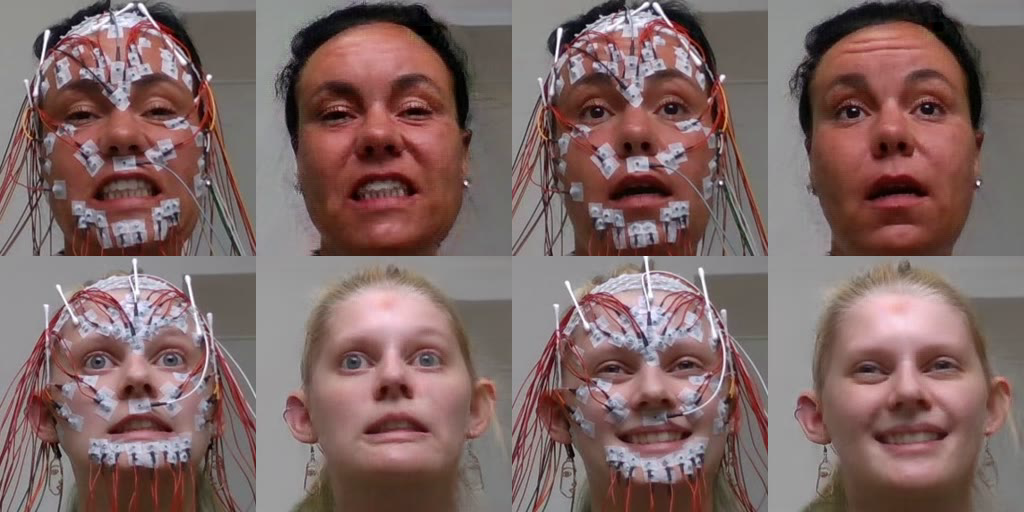}
  \caption{
    Overview of two test subjects with their respective sEMG sensor removal.
    The covered facial features were restored in all examples.
  }
  \label{fig:comp}
\end{figure}

We display a selection of the resulting pairs for the sensor removal in Fig.~\ref{fig:comp}.
Additionally, in the supplementary material we also provide the entire videos of the~\sensor~($S$) and~\clean~($C$) videos side-by-side for better inspection.
The visual results already indicate a correct restoration of the underlying facial features.
We assume that the model learned a generalized version of each test subject's face as in some of the shown examples the view was zoomed out.
Thus, missing information must have been encoded inside the model.
The examples show that the model retains head posture, orientation, and most significantly the correct facial expressions.
However, to ensure that no underlying artifacts are introduced by the generation process, we evaluate the images with existing state-of-the-art models on their correct results.
For this quantitative evaluation, we compute four scores per session, totaling to nine measurements per individual test subject including the baseline.

\subsection{Perceptual Similarity Comparison}
\begin{table*}[t]
    \centering
    \caption{The perceptual scores computed over all test subjects indicate that the clean videos resemble the normal videos more than the sensor videos. One can even see that some reconstruction yield even better results than our baseline. }
    \label{tab:perception}
    \begin{tabularx}{\linewidth}{YRRRRRR}
        \toprule
        {}            & \multicolumn{3}{c}{LPIPS}      & \multicolumn{3}{c}{FID}                                                                                                                                           \\
        {}            & Schaede                        & Emotion                       & Sentence                       & Schaede                        & Emotion                        & Sentence                       \\
        \midrule
        $N_{1}-N_{2}$ & \texttt{0.27\tiny{$\pm$0.09}}  & \texttt{0.26\tiny{$\pm$0.07}} & \texttt{0.27\tiny{$\pm$0.09}}  & \texttt{68.6\tiny{$\pm$30.7}}  & \texttt{72.6\tiny{$\pm$33.2}}  & \texttt{65.5\tiny{$\pm$34.5}}  \\
        $N_{1}-S_{1}$ & \texttt{0.55\tiny{$\pm$0.04}}  & \texttt{0.56\tiny{$\pm$0.05}} & \texttt{0.56\tiny{$\pm$0.06}}  & \texttt{278.3\tiny{$\pm$28.1}} & \texttt{280.0\tiny{$\pm$26.9}} & \texttt{279.0\tiny{$\pm$32.3}} \\
        $N_{1}-C_{1}$ & \texttt{0.25\tiny{$\pm$0.09}}  & \texttt{0.26\tiny{$\pm$0.11}} & \texttt{0.27\tiny{$\pm$0.12}}  & \texttt{48.4\tiny{$\pm$17.2}}  & \texttt{50.8\tiny{$\pm$20.5}}  & \texttt{54.4\tiny{$\pm$44.2}}  \\
        $N_{2}-S_{2}$ & \texttt{0.55\tiny{$\pm$0.03}}  & \texttt{0.55\tiny{$\pm$0.04}} & \texttt{0.55\tiny{$\pm$0.03}}  & \texttt{279.8\tiny{$\pm$29.3}} & \texttt{278.6\tiny{$\pm$27.3}} & \texttt{277.0\tiny{$\pm$32.9}} \\
        $N_{2}-C_{2}$ & \texttt{0.26\tiny{$\pm$0.01 }} & \texttt{0.25\tiny{$\pm$0.09}} & \texttt{0.27\tiny{$\pm$0.01 }} & \texttt{59.3\tiny{$\pm$34.6}}  & \texttt{61.5\tiny{$\pm$34.2}}  & \texttt{57.9\tiny{$\pm$37.5}}  \\
        \bottomrule
    \end{tabularx}
\end{table*}

To ensure that the qualitative results are not only visually correct, we analyze the frames of the videos and compute a perceptual score.
The comparison between the two~\normal~videos ($N_{1}$, $N_{2}$) represents a perfect match.
This will be the baseline for all our evaluations.
The score of a certain combination is the average over all respective videos.
Furthermore, we investigate the results of each task separately.
Table~\ref{tab:perception} displays the results for the LPIPS and FID scores.
We show the mean scores over all test subjects with their respective standard deviation.
The generated~\clean~videos $C_{1}$ and $C_{2}$ have a considerably higher resemblance to the baseline than the~\sensor~videos $S_{1}$ and $S_{2}$.
This is a strong indicator that our reconstruction method works correctly.
Furthermore, the LPIPS scores indicate that there is only little difference between the three given tasks.
However, the results of the FID score yield different results in that regard.
As for the FID $128$ images are considered to estimate the underlying image generative distribution, the different facial movements could be the reason for the differences among the tasks.
Another interesting observation is that the scores of all the reconstructed video comparisons yield better results than our baseline.
We assume that the differences in the background, due to changes of the recording setup between the sessions, could be the major contributing factor.
Together with the images, as seen in Fig.~\ref{fig:experimental_setup} and Fig.~\ref{fig:comp}, the perceptual scores indicate that our generator network correctly recreates the faces of the test subjects.
We provide the side-by-side emotional task of the~\sensor~and~\clean~videos in the supplementary material. 
These will show that the underlying facial features are reconstructed correctly.
One can see that even small facial movements, including eyelid closing and gazes, are correctly restored.

\subsection{Action Unit Reconstruction}
We compare the reconstruction of AUs using two feature extraction methods.
The first method is a random decision forest (RDF)~\cite{breimanRandomForests2001} trained on the 68 facial landmarks extracted with MobileNet by Howard et al.~\cite{howardMobileNetsEfficientConvolutional2017}.
The second model JAA-NET by Shao et al.~\cite{shaoJAANetJointFacial2021} includes a custom landmark detector and AUs estimator.
The results in Table~\ref{tab:aus} show the results for both methods including the similarity between time series compared with dynamic time warping (DTW)~\cite{lhermitteComparisonTimeSeries2011} and mean absolute percentage error (MAPE).
For the comparison we average the results of test subjects and AUs and show their respective mean scores and deviation.
We separate the results into the three tasks to investigate differences among different facial movements.
Please note that the RDF model computes 20 and the JAA-NET model only 12 AUs.
Thus a comparison between these models is not fully possible.
Furthermore, we exclude \texttt{AU43} (eye blinking~\cite{ekmanFacialActionCoding1978}) from our analysis as it differs in all videos.
\begin{figure}[!ht]
  \centering
  \includegraphics[width=\linewidth]{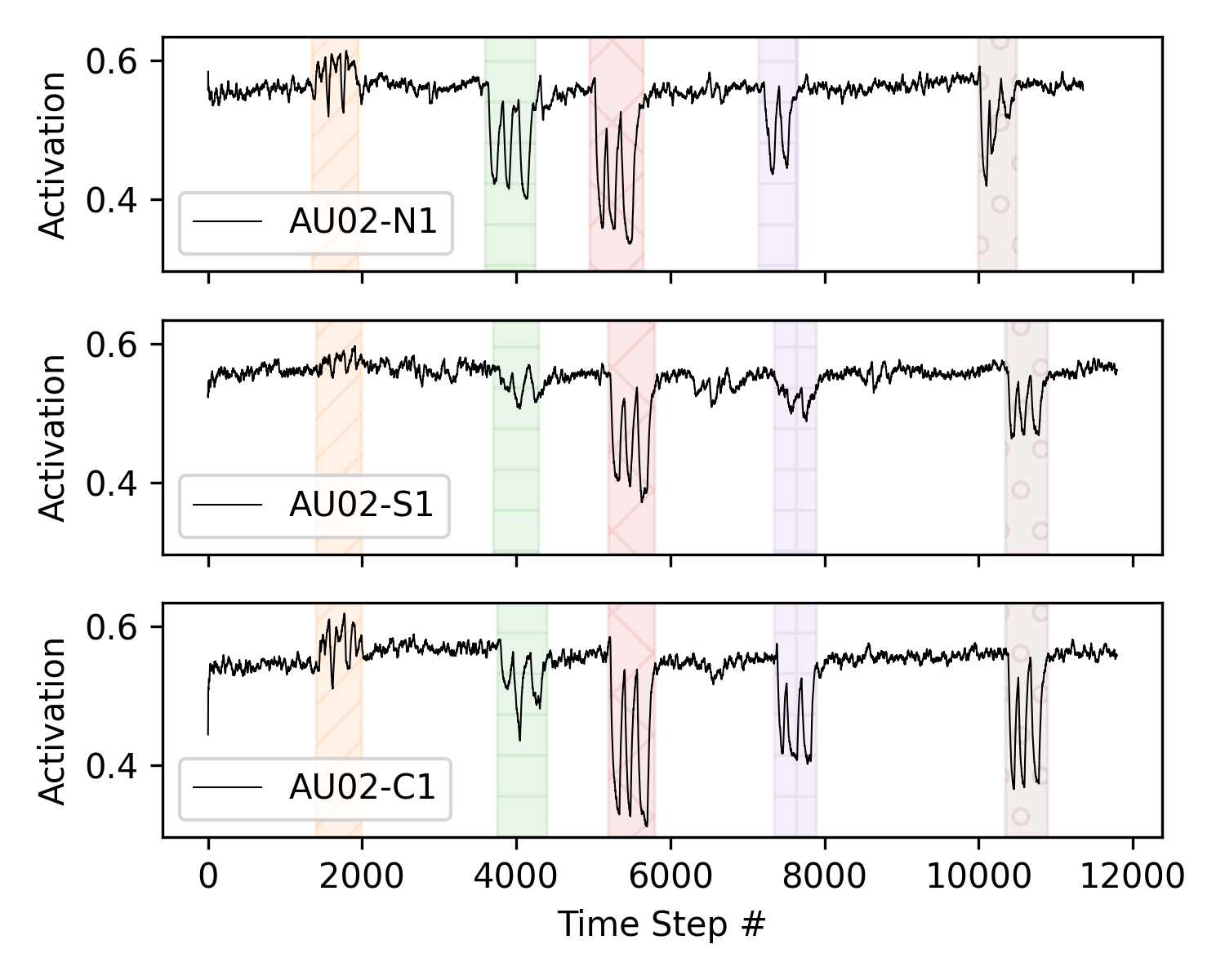}
  \caption{
    Qualitative comparison between the \texttt{AU02} and the restoration of the interesting intervals during the Schaede task using the RDF model.
    We highlight five intervals of activation, which should be detected.
    Our approach can restore missing intervals and correct the amplitude of existing ones.
  }
  \label{fig:aus_rec}
\end{figure}

\begin{table*}[t]
        \centering
        \caption{The similarity between the Action Unit time series is computed. The results indicate the AUs can be computed correctly in \textit{clean} videos, whereas the \textit{sensor} videos yield wrong results. This is even more evident for the results of the JAA-NET model.}
        \label{tab:aus}
        \begin{tabularx}{\textwidth}{ccRRRRRR}
                \toprule
                                                                                           &               & \multicolumn{3}{c}{DTW}       & \multicolumn{3}{c}{MAPE}                                                                                                                                              \\
                                                                                           &               & Schaede                       & Emotion                       & Sentence                      & Schaede                         & Emotion                         & Sentence                          \\
                \midrule
                \parbox[t]{2mm}{\multirow{5}{*}{\rotatebox[origin=c]{90}{\tiny{RDF}}}}     & $N_{1}-N_{2}$ & \texttt{2.1\tiny{$\pm$1.1}}   & \texttt{2.0\tiny{$\pm$1.0}}   & \texttt{1.4\tiny{$\pm$0.8}}   & \texttt{0.09\tiny{$\pm$0.05}}   & \texttt{0.11\tiny{$\pm$0.05}}   & \texttt{0.09\tiny{$\pm$0.05}}     \\
                                                                                           & $N_{1}-S_{1}$ & \texttt{2.8\tiny{$\pm$1.4}}   & \texttt{2.4\tiny{$\pm$1.2}}   & \texttt{1.7\tiny{$\pm$0.9}}   & \texttt{0.11\tiny{$\pm$0.05}}   & \texttt{0.12\tiny{$\pm$0.05}}   & \texttt{0.10\tiny{$\pm$0.05}}     \\
                                                                                           & $N_{1}-C_{1}$ & \texttt{2.3\tiny{$\pm$1.2}}   & \texttt{1.9\tiny{$\pm$1.0}}   & \texttt{1.4\tiny{$\pm$0.8}}   & \texttt{0.09\tiny{$\pm$0.05}}   & \texttt{0.10\tiny{$\pm$0.05}}   & \texttt{0.09\tiny{$\pm$0.04}}     \\
                                                                                           & $N_{2}-S_{2}$ & \texttt{2.8\tiny{$\pm$1.4}}   & \texttt{2.4\tiny{$\pm$1.3}}   & \texttt{1.7\tiny{$\pm$0.9}}   & \texttt{0.11\tiny{$\pm$0.05}}   & \texttt{0.12\tiny{$\pm$0.05}}   & \texttt{0.10\tiny{$\pm$0.05}}     \\
                                                                                           & $N_{2}-C_{2}$ & \texttt{2.4\tiny{$\pm$1.2}}   & \texttt{2.2\tiny{$\pm$1.1}}   & \texttt{1.3\tiny{$\pm$0.6}}   & \texttt{0.10\tiny{$\pm$0.05}}   & \texttt{0.11\tiny{$\pm$0.05}}   & \texttt{0.08\tiny{$\pm$0.04}}     \\
                \cline{1-8}
                \parbox[t]{2mm}{\multirow{5}{*}{\rotatebox[origin=c]{90}{\tiny{JAA-NET}}}} & $N_{1}-N_{2}$ & \texttt{~5.1\tiny{$\pm$~3.3}} & \texttt{~3.8\tiny{$\pm$~2.1}} & \texttt{~3.0\tiny{$\pm$~2.0}} & \texttt{~2.56\tiny{$\pm$~3.73}} & \texttt{~1.43\tiny{$\pm$~0.79}} & \texttt{~~1.20\tiny{$\pm$~~0.70}} \\
                                                                                           & $N_{1}-S_{1}$ & \texttt{25.6\tiny{$\pm$24.9}} & \texttt{16.6\tiny{$\pm$16.1}} & \texttt{17.0\tiny{$\pm$16.5}} & \texttt{13.15\tiny{$\pm$16.03}} & \texttt{10.12\tiny{$\pm$11.57}} & \texttt{~23.09\tiny{$\pm$~28.71}} \\
                                                                                           & $N_{1}-C_{1}$ & \texttt{~4.6\tiny{$\pm$~2.9}} & \texttt{~3.6\tiny{$\pm$~2.1}} & \texttt{~3.0\tiny{$\pm$~2.1}} & \texttt{~1.75\tiny{$\pm$~1.98}} & \texttt{~1.57\tiny{$\pm$~1.37}} & \texttt{~36.13\tiny{$\pm$109.88}} \\
                                                                                           & $N_{2}-S_{2}$ & \texttt{24.0\tiny{$\pm$23.5}} & \texttt{15.8\tiny{$\pm$15.6}} & \texttt{16.4\tiny{$\pm$16.2}} & \texttt{10.54\tiny{$\pm$11.20}} & \texttt{10.46\tiny{$\pm$10.53}} & \texttt{~17.39\tiny{$\pm$~19.45}} \\
                                                                                           & $N_{2}-C_{2}$ & \texttt{~4.5\tiny{$\pm$~2.8}} & \texttt{~3.7\tiny{$\pm$~2.0}} & \texttt{~2.8\tiny{$\pm$~1.8}} & \texttt{~2.38\tiny{$\pm$~4.43}} & \texttt{29.80\tiny{$\pm$98.40}} & \texttt{~~2.90\tiny{$\pm$~~4.40}} \\
                \bottomrule
        \end{tabularx}
\end{table*}

We can see that for most~\clean~videos we reach a similar score as the baseline.
The qualitative reconstruction of the correct feature extraction can be seen in Fig.~\ref{fig:aus_rec}.
We highlight the time series for \texttt{AU02} (outer brow raiser~\cite{ekmanFacialActionCoding1978}) during the Schaede~\cite{schaedeVideoInstructionSynchronous2017} video estimated by the RDF model.
During the video five distinct areas can be detected inside the~\normal~videos.
However, in the~\sensor~these five areas are either not distinguishable from noise or detected with a wrong intensity.
The time series for~\clean~video however shows that all five area could be restored including a more similar intensity to the~\normal~videos.
In the supplementary material we provide further qualitative comparisons to indicate the robustness of the restoration.
However, it can also be seen in Table~\ref{tab:emotions} that the restoration scores for the Emotion and Sentence task do not reach the baseline scores.
As these videos are at the end of each recording session, we assume that the test subjects differ significantly in their behavior and the MAPE metric cannot handle this differences in the time series.
Therefore, we can also conclude that the DTW comparison is a more stable metric to compare the restoration of the obstructed facial features.

\subsection{Emotion Detection Comparison}

\begin{table*}[t]
    \centering
    \caption{The reconstruction of the obstructed facial features in regards to the emotional states achieves similar scores to the baseline. There are no real differences evident between the three different tasks.}
    \label{tab:emotions}
    \begin{tabularx}{\textwidth}{YRRRRRR}
        \toprule
        {}            & \multicolumn{3}{c}{DTW}         & \multicolumn{3}{c}{MAPE}                                                                                                                                                \\
        {}            & Schaede                         & Emotion                         & Sentence                        & Schaede                         & Emotion                         & Sentence                        \\
        \midrule
        $N_{1}-N_{2}$ & \texttt{~~5.0\tiny{$\pm$~~2.4}} & \texttt{~~4.6\tiny{$\pm$~~0.9}} & \texttt{~~2.4\tiny{$\pm$~~1.7}} & \texttt{~~4.5\tiny{$\pm$~~3.0}} & \texttt{~~5.5\tiny{$\pm$~~3.2}} & \texttt{~~1.9\tiny{$\pm$~~0.8}} \\
        $N_{1}-S_{1}$ & \texttt{~19.3\tiny{$\pm$~16.7}} & \texttt{~13.8\tiny{$\pm$~~8.1}} & \texttt{~12.3\tiny{$\pm$~14.4}} & \texttt{~57.5\tiny{$\pm$106.9}} & \texttt{~30.8\tiny{$\pm$~50.7}} & \texttt{~36.9\tiny{$\pm$~69.5}} \\
        $N_{1}-C_{1}$ & \texttt{~~6.3\tiny{$\pm$~~3.5}} & \texttt{~~5.9\tiny{$\pm$~~1.7}} & \texttt{~~3.5\tiny{$\pm$~~2.9}} & \texttt{~~8.5\tiny{$\pm$~~7.9}} & \texttt{~~6.5\tiny{$\pm$~~3.7}} & \texttt{~~6.3\tiny{$\pm$~~4.9}} \\
        $N_{2}-S_{2}$ & \texttt{~18.5\tiny{$\pm$~14.8}} & \texttt{~12.9\tiny{$\pm$~~6.8}} & \texttt{~11.2\tiny{$\pm$~13.7}} & \texttt{~45.3\tiny{$\pm$~67.4}} & \texttt{~20.8\tiny{$\pm$~29.3}} & \texttt{~34.9\tiny{$\pm$~58.2}} \\
        $N_{2}-C_{2}$ & \texttt{~~6.6\tiny{$\pm$~~3.5}} & \texttt{~~6.0\tiny{$\pm$~~1.6}} & \texttt{~~3.3\tiny{$\pm$~~2.7}} & \texttt{~~8.8\tiny{$\pm$~10.1}} & \texttt{~~4.8\tiny{$\pm$~~2.7}} & \texttt{~~6.1\tiny{$\pm$~~6.5}} \\
        \bottomrule
    \end{tabularx}
\end{table*}

\begin{figure}[t]
  \centering
  \includegraphics[width=\linewidth]{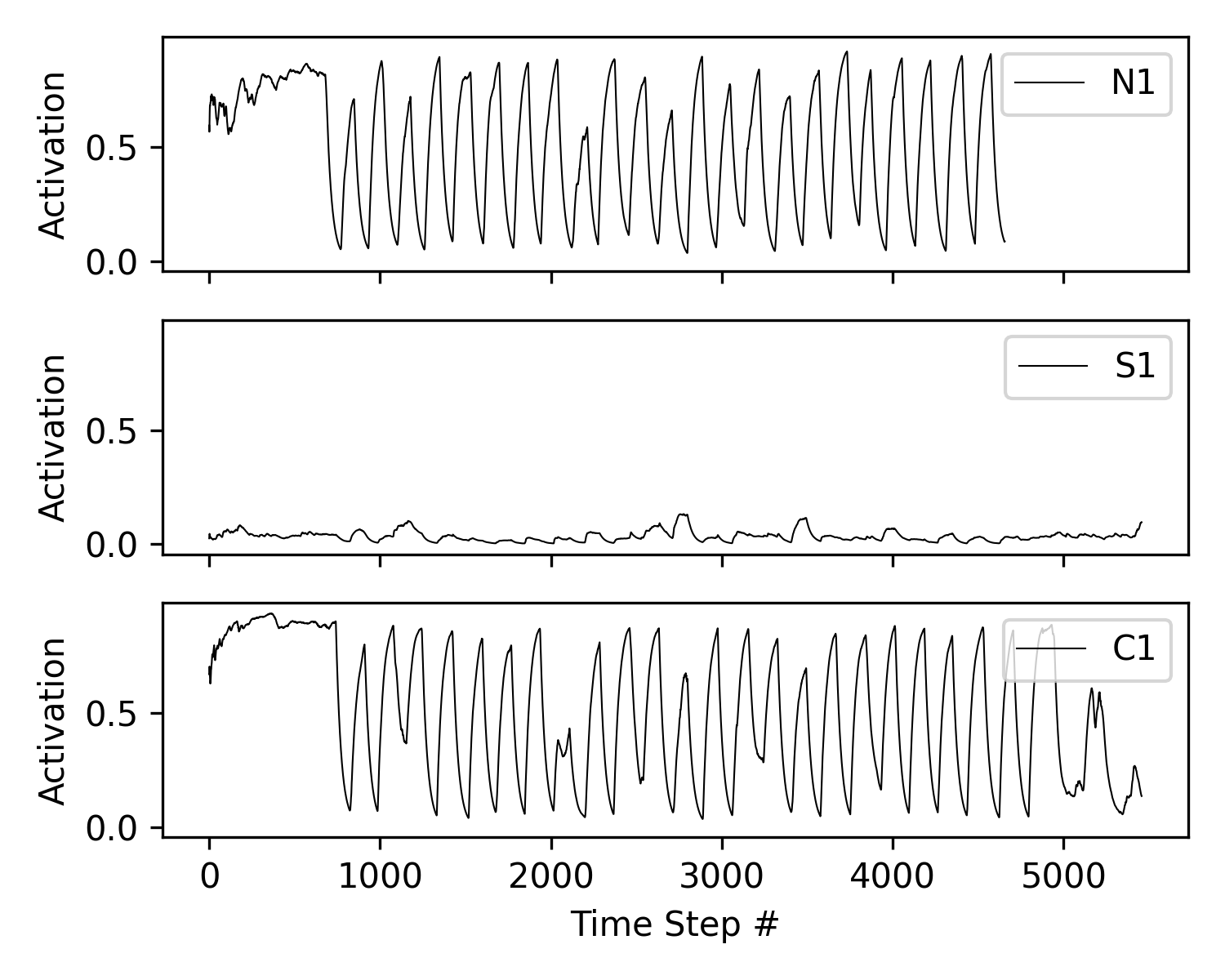}
  \caption{
    The qualitative results show that the time series for the \texttt{neutral} emotional state can be correctly restored.
    The oscillating pattern between the neutral and other emotional states does not appear inside the~\sensor~video.
    After removing the sEMG sensors in the test subject's face the ResMaskNet can correctly estimate the facial appearance.
  }
  \label{fig:emo_rec}
\end{figure}

We evaluate the emotion detection restoration on all three tasks, whereas the Emotion task should weight most in the quantitative analysis.
To create the time series for each of the seven basic emotions, we use the ResMaskNet by Luan et al.~\cite{luanresmaskingnet2020} on each single video frame.
Then we compare the given video pairs using DTW and MAPE.
In Table~\ref{tab:emotions} we show the results over all 36 test subjects averaged over all emotions.
The table contains the mean distance and deviation for all three tasks.
It can can be seen that our restoration method achieves similar scores to our baseline evaluation.
In Fig.~\ref{fig:emo_rec} we display the reconstruction of the time series for the \texttt{neutral} emotional state.
In the~\sensor~video the ResMaskNet cannot detect any correct \texttt{neutral} state of the test subject.
As the test subjects constantly switch between emotions back to the \texttt{neutral} an oscillating pattern should be visible.
This is the case for the~\normal~videos and our~\clean~video time series.
However, in the~\sensor~video this pattern does not appear.

\section{\uppercase{Limitations}}
\begin{figure}[t]
  \centering
  \includegraphics[width=\linewidth,trim={0 5.2cm 0 0},clip]{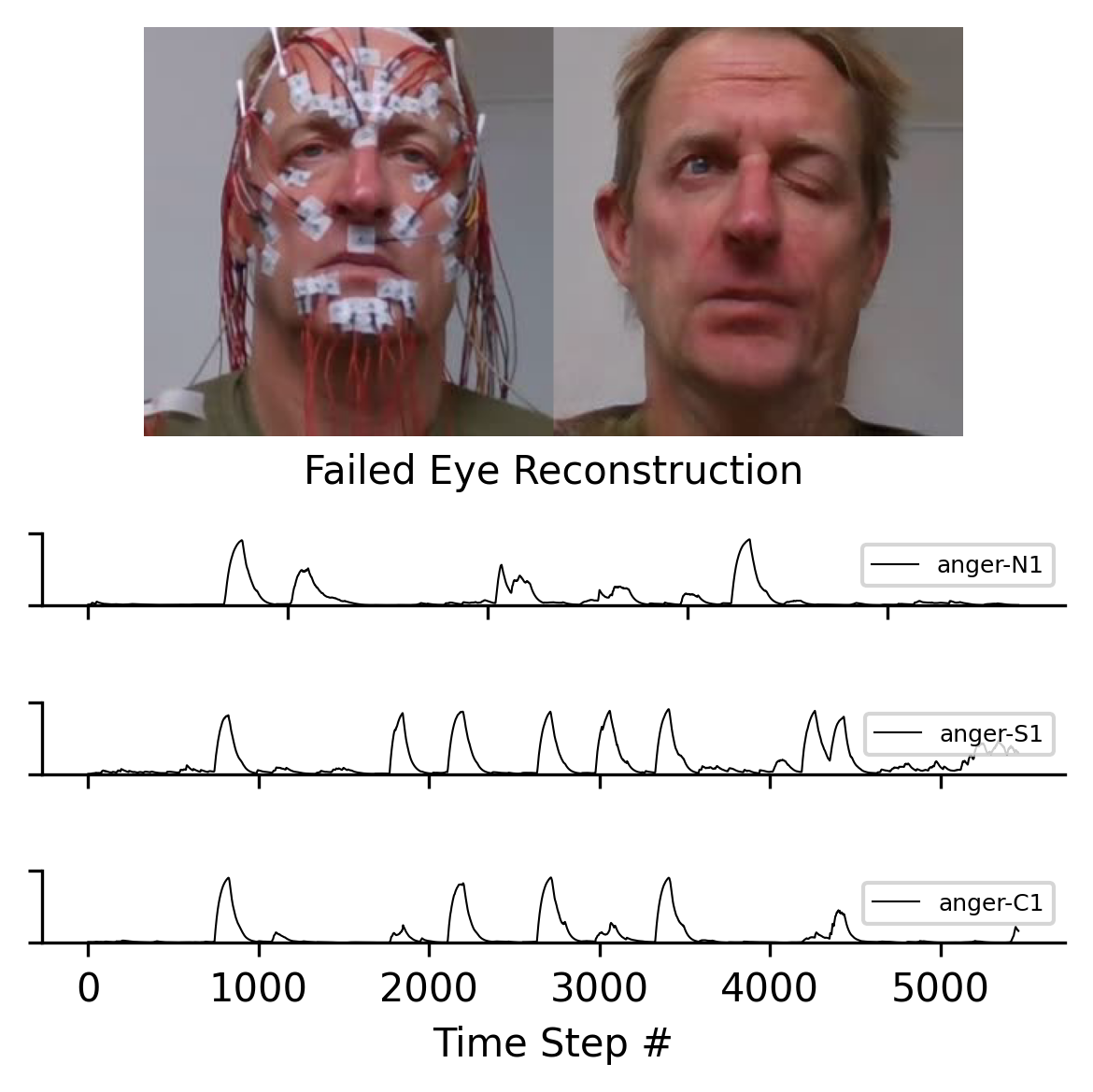}
  \caption{
    There are some limitations to our approach for obstructed facial feature reconstruction.
    For instance, during the sEMG sensor removal the model closed the left eye of a test subject.
  }
  \label{fig:lim_eye}
\end{figure}

\begin{figure}[h]
  \centering
  \includegraphics[width=\linewidth,trim={0 0 0 4.5cm},clip]{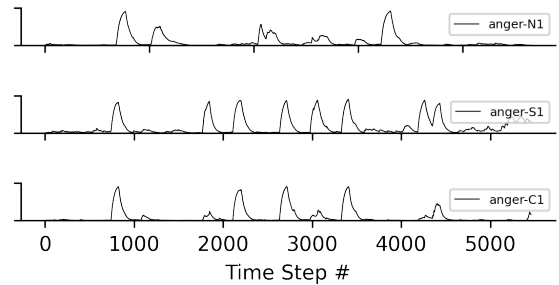}
  \caption{
    The sEMG sensors do not obstruct the detection of the \texttt{anger} emotion with ResMaskNet.
  }
  \label{fig:lim_emo}
\end{figure}

The correctness of our proposed method for restoring facial features is depending on the quality of the selected frames during training.
In our ablation studies we observed the following limitations. 
When selecting frames equidistantly, there was a chance that eye movement and eyelid closing was not represented in the data.
Thus, the model was not able to restore these movements in the resulting full video.
We found that a random selection of frames mitigated this undesirable behavior.
Additionally, we observed that dynamically computing the bounding box of the face required more training time.
Face size and visible areas vary a lot during a sequence and the face detector does not always yield a perfect fit.
Furthermore, sometimes artifacts are introduced in the generation process as seen in Fig.~\ref{fig:lim_eye}, which can be attributed to the selection of training frames.
We also observed that the recognition of the \texttt{anger} emotion using ResMaskNet was less affected by sEMG sensors in comparison to other emotions, as can be seen in Fig.~\ref{fig:emo_rec} and \ref{fig:lim_emo}.
Relevant facial parts of the \texttt{anger} emotion like the mouth area are not obstructed.
It is worth noting, that our sensor removal method could still slightly improve such scenarios. 

\section{\uppercase{Conclusion}}
In this paper, we demonstrated that instead of fine-tuning models to fit obstructed unlabeled data, we can correctly restore previously hidden facial features using a CycleGAN~\cite{zhuUnpairedImagetoImageTranslation2017} approach.
We evaluated our method with respect to the visual quality of the generated faces using perceptual scores.
Evaluation has been carried out by looking into pairwise frame similarity and by estimating the underlying image distributions.
Both methods clearly indicated a high visual similarity of the~\clean~videos (reconstruction) and the~\normal~videos (no obstructions).
Further, we investigated the correctness of state-of-the-art facial analysis methods based on these reconstructed facial features.
Here, the results showed that restoration quality is good enough to successfully apply methods for Facial Actions Units~\cite{ekmanFacialActionCoding1978} and emotions detection afterwards.
In our work specifically, this allowed us to further progress in the area of connecting mimics and underlying facial muscles, as existing vision-based methods can now be applied directly.
Furthermore, the data driven approach based on individual test subjects makes this approach applicable to any person disregarding age, gender, and ethnicity.

\vfill
\section*{\uppercase{Acknowledgements}}
This work has been funded by the Deutsche Forschungsgemeinschaft (DFG - German Research Foundation) project 427899908 \textit{BRIDGING THE GAP: MIMICS AND MUSCLES}.

\bibliographystyle{apalike}
{\small
  \bibliography{visapp_paper}}

\end{document}